\documentclass{article}

\usepackage{arxiv}

\usepackage[utf8]{inputenc} % allow utf-8 input
\usepackage[T1]{fontenc}    % use 8-bit T1 fonts
\usepackage{hyperref}       % hyperlinks
\usepackage{url}            % simple URL typesetting
\usepackage{booktabs}       % professional-quality tables
\usepackage{amsfonts}       % blackboard math symbols
\usepackage{nicefrac}       % compact symbols for 1/2, etc.
\usepackage{microtype}      % microtypography
\usepackage{lipsum}		% Can be removed after putting your text content
\usepackage{graphicx}
\usepackage{natbib}
\usepackage{doi}

\usepackage{algorithm}
\usepackage{algorithmic}
\usepackage{amsmath}
\usepackage{booktabs}

\usepackage{amsfonts}
\usepackage{graphicx}
\usepackage{float}
\usepackage{subcaption}
\usepackage{pgfplots}

\title{
A new membership inference attack that spots memorization in generative and predictive models: Loss-Based with Reference Model algorithm (LBRM)}

\author{ \href{https://orcid.org/0000-0000-0000-0000}{\includegraphics[scale=0.06]{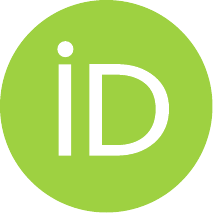}\hspace{1mm} Faiz TALEB}\\
	EDF, Samovar, T\'el\'ecom SudParis, Institut Polytechnique de Paris\\
	\texttt{faiz.taleb@edf.fr} \\
	\And
	\href{https://orcid.org/0000-0000-0000-0000}{\includegraphics[scale=0.06]{orcid.pdf}\hspace{1mm}Ivan GAZEAU} \\
	EDF \\
	\texttt{ivan.gazeau@edf.fr} \\
	\AND
        \href{https://orcid.org/0000-0000-0000-0000}{\includegraphics[scale=0.06]{orcid.pdf}\hspace{1mm} Maryline  LAURENT}\\
        Samovar, T\'el\'ecom SudParis, Institut Polytechnique de Paris\\
        \texttt{Maryline.Laurent@telecom-sudparis.eu} \\
 }
\renewcommand{\shorttitle}{\textit{LRBM} approach for MIA}

\hypersetup{
pdftitle={A template for the arxiv style},
pdfsubject={q-bio.NC, q-bio.QM},
pdfauthor={David S.~Hippocampus, Elias D.~Striatum},
pdfkeywords={First keyword, Second keyword, More},
}

\newcommand{\targetmodel}{$\mathcal{T}$ }
\newcommand{\referencemodel}{$\mathcal{R}$ }

\newcommand{\publicdataset}{$\mathcal{O}$ }
\newcommand{\privatedataset}{$\mathcal{P}$ }
\newcommand{\threshold}{$\theta $}

\begin{document}
\maketitle

\begin{abstract}
	Generative models can unintentionally memorize training data, posing significant privacy risks. This paper addresses the memorization phenomenon in time series imputation models, introducing the Loss-Based with Reference Model (LBRM) algorithm. The LBRM method leverages a reference model to enhance the accuracy of membership inference attacks, distinguishing between training and test data. Our contributions are twofold: first, we propose an innovative method to effectively extract and identify memorized training data, significantly improving detection accuracy. On average, without fine-tuning, the AUROC improved by approximately 40\%. With fine-tuning, the AUROC increased by approximately 60\%. Second, we validate our approach through membership inference attacks on two types of architectures designed for time series imputation, demonstrating the robustness and versatility of the LBRM approach in different contexts. These results highlight the significant enhancement in detection accuracy provided by the LBRM approach, addressing privacy risks in time series imputation models.
\end{abstract}

% keywords can be removed

\section{Introduction}

The memorization problem in generative AI has been a topic of interest in the literature. Carlini et al. \cite{carlini_extracting_nodate} first introduced the concept in the context of LLM models, highlighting how GPT-2 unintentionally memorizes small data. Since then, numerous studies have been devoted to this problem. This phenomenon later came to be known as Unintended Memorization in \cite{carlini_secret_2019}. Memorization problems pose a significant privacy and confidentiality concern, because generative models could potentially memorize sensitive data, which a malicious user could exploit to extract sensitive information from the model.

Privacy attacks aim at extracting sensitive information from the generative model. Among the various privacy attacks, inference attacks such as Membership Inference Attack (MIA) or attribute inference are notable. The goal is not necessarily to extract data from the model, but to infer sensitive information. For example, in membership inference attacks, the attacker determines whether a specific row of data was used in the training of the deep model. Since membership attacks are often easier and require less effort, numerous examples can be found in the literature \cite{liu_encodermi_2021,fu_practical_2023,wu_adapting_2021,he_node-level_2021,carlini_membership_2022,webster_this_2021,olatunji_membership_2021,zhang_generated_2024}.

The imputation of missing values in multivariate time series is essential to ensure the accuracy and reliability of subsequent analysis and decision-making processes. Multivariate time series data in various domains, such as IoT systems, clinical trials, and financial and transportation systems, often suffer from missing values due to sensor failures, unstable environments, irregular sampling, privacy concerns, system downtime, and communication issues. These missing values can significantly affect the quality of downstream analysis. Therefore, developing effective imputation methods, especially those that leverage advanced deep learning architectures, is essential for maintaining data integrity and improving the performance of subsequent analyses.

As it's mentioned in \cite{wang_deep_2025}, there are two main categories of methods for deep learning-based time series imputation: predictive methods and generative methods. Predictive methods include techniques such as Recurrent Neural Networks (RNNs), Convolutional Neural Networks (CNNs), and attention-based models. Generative methods include approaches such as  Autoencoders (AEs), Generative Adversarial Networks (GANs), and diffusion models. 

In this work, we investigated the risk of unintended memorization in time series imputation models. We show that time series models are also susceptible to information leakage during their deployment. We propose a method that can extract and identify whether the extracted data was part of the training dataset. Our approach can be used in a completely black-box setting with query access only.
A significant challenge in time series memorization is distinguishing whether a prediction is a genuine forecast or a memorized output of the model. To determine whether the extracted data is real, we use what we call a reference model. The goal of the witness model is to provide an unbiased reference that helps determine whether the data represents a good prediction or memorization.
Our attack methodology is applicable to all types of generative or predictive time series models. In our study, we evaluated the effectiveness of our attack on two different architectures designed for time series imputation: one predictive based on attention mechanisms and the other generative based on autoencoders.
In summary, this paper makes two major contributions:
\begin{itemize}
    \item A novel approach to effectively extract and identify memorized training data in time series imputation: We introduce the Loss-Based with Reference Model (LBRM) algorithm, which leverages a reference model to enhance the accuracy of distinguishing between training and test data.
    \item Validation of our approach through membership inference attacks on two types of architectures designed for time series imputation: We evaluated the effectiveness of our attack on a predictive model based on attention mechanisms (SAITS) and a generative model based on autoencoders (AE). Our results demonstrate the robustness and versatility of the LBRM approach in different contexts, highlighting its potential to address privacy risks.
\end{itemize}

\section{Related Work}
Generative AI has received considerable attention in recent years, with applications in diverse domains such as text generation, image synthesis, and music composition. Some popular large language models (LLMs) include GPT, Llama, and Mistral, which are based on transformer architectures \cite{openai_gpt-4_2024,touvron_llama_2023,jiang_mistral_2023}. 

The memorization problem in generative AI has been a topic of interest in the literature. Carlini et al. \cite{carlini_extracting_nodate} first introduced the concept in the context of LLMs models, highlighting how GPT-2 unintentionally memorizes small data. Since then, numerous studies have been devoted to this problem. This phenomenon later became known as Unintended Memorization in \cite{carlini_secret_2019}. Memorization problems pose a significant privacy and confidentiality concern, as generative models could potentially memorize sensitive data that a malicious user could exploit to extract sensitive information from the model.

 Among the various privacy attacks, inference attacks such as membership inference or attribute inference are notable. The goal is not necessarily to extract data from the model, but to infer sensitive information. For example, in membership inference attacks, the attacker determines whether a particular row of data was used in training the deep model. Because membership attacks are often simpler and require less effort, there are many examples in the literature \cite{liu_membership_2022,hilprecht_monte_2019,hilprecht_reconstruction_2019,fu_practical_2023,wu_adapting_2021,he_node-level_2021,carlini_membership_2022,webster_this_2021,olatunji_membership_2021,hayes_logan_2018,zhang_generated_2024}. 
 
 One particular work is by Liu et al. \cite{liu_membership_2022}, which proposes a novel membership inference attack method called TRAJECTORYMIA. This method leverages the loss trajectory information from the training process of the target model to improve the performance of membership inference attacks. The key idea is to utilize the sequence of losses evaluated on the target model at different training epochs, known as the loss trajectory, to distinguish between member and non-member samples. They employ knowledge distillation to obtain the loss trajectory in practical scenarios where only the final trained model is accessible. The attack model then uses the distilled loss trajectory and the loss of the final model to infer membership status.

Another notable work is by Fu et al. \cite{fu_practical_2023}, which introduces SPV-MIA (Self-calibrated Probabilistic Variation Membership Inference Attack). This method aims to infer membership in fine-tuned large language models (LLMs) by detecting memorization rather than overfitting. Fu et al. propose using probabilistic variation as a membership signal, which is assessed by a paraphrasing model and calibrated by a self-prompt reference model. The self-prompt approach involves generating a reference dataset by prompting the target LLM itself, allowing the adversary to collect a dataset with a similar distribution from public APIs.

The main difference between our approach and the methods \cite{liu_membership_2022} and \cite{fu_practical_2023} lies in the application and improvement of membership inference attacks. While \cite{liu_membership_2022} focuses on exploiting loss trajectory information for machine learning classifiers and \cite{fu_practical_2023} uses probabilistic variation to detect memorization in fine-tuned large language models, our research extends these concepts to time series imputation. We incorporate reference models to improve accuracy and address the unique challenges of time series data, applying our method in both fine-tuned and unfine-tuned scenarios. Additionally, in the context of time series imputation, we have applied our approach to two different types of architectures: one based on transformers and the other based on autoencoders. This dual approach allows us to provide a robust solution for inferring membership status in time series data.

%plutot pour l'intro

\section{Background}
We refer to the model under attack as the target model. In this paper, we start by building two models for time series imputation from scratch. In our study, there are two parts, the first is to assess the unintended memorization in time series imputation by doing a membership inference attack.

\subsection{Definition of Multivariate Time Series Imputation}
We stick to the definition found in \cite{du_saits_2023}. \\Given a collection of multivariate time series with $T$ time steps and $D$ dimensions, it is denoted as $\mathbf{X} = \{\mathbf{x}_1, \mathbf{x}_2, \ldots, \mathbf{x}_t, \ldots, \mathbf{x}_T\} \in \mathbb{R}^{T \times D}$, where the $t$-th step $\mathbf{x}_t = \{x_{1t}, x_{2t}, \ldots, x_{dt}, \ldots, x_{Dt}\} \in \mathbb{R}^{1 \times D}$ and any value in it may be missing. Similarly, $X_{dt}$ represents the $d$-th dimension variable of the $t$-th step in $\mathbf{X}$. To represent the missing variables in $\mathbf{X}$, the missing mask vector $\mathbf{M} \in \mathbb{R}^{T \times D}$ is introduced, where:

\[
M_{dt} =
\begin{cases}
1 & \text{if } X_{dt} \text{ is observed} \\
0 & \text{if } X_{dt} \text{ is missing}
\end{cases}
\]

\subsection{Membership attack}
Let $\mathcal{D}$ be the training dataset and $\mathcal{M}$ be a machine learning model trained on $\mathcal{D}$. Let $\mathbf{x}$ be a data point that may or may not be in $\mathcal{D}$. The goal of a membership attack is to determine whether $\mathbf{x} \in \mathcal{D}$.

Formally, a membership attack can be defined as the following attack function: \\
%$\mathcal{A}: \mathcal{X} \times \mathcal{M} \to \{0, 1\}$
\[
\mathcal{A}(\mathbf{x}, \mathcal{M}) =
\begin{cases}
1 & \text{if } \mathbf{x} \in \mathcal{D} \\
0 & \text{if } \mathbf{x} \notin \mathcal{D}
\end{cases}
\]
where $\mathbf{x}  \in \mathcal{X}$ and $\mathcal{X}$ is the space of possible data points.

\section{Exploiting Memorization for Membership Inference Attacks with our LBRM Approach}
% Dire que nous essayons d'exploiter la faille de mémorization pour faire une MIA
% L'idée est de d'essayer de faire une MIA sur des données qui ont été mémorisées par le modèle préalablement
In this section, we describe our approach to performing a membership inference attack (MIA) on generative or predictive models. We introduce our Loss-Based with Reference Model (LBRM) algorithm, which facilitates MIA on data memorized by the target model.

Our designed \textit{Loss-Based with Reference Model algorithm (LBRM)}  aims to exploit the phenomenon of memorization by the model during the training process to perform a MIA. This attack is performed by comparing the performance of a target model (Model \targetmodel) with that of a reference model (Model \referencemodel). The algorithm takes as input the Model \targetmodel, to be attacked, the Model \referencemodel, the suspicious data $x$ and a threshold \threshold. As output, it returns the classification of the data $x$ as member or not. The LBRM algorithm is detailed in Algorithm \ref{Algo1}.

\begin{algorithm}[H]
\caption{Loss-Based with Reference Model (LBRM) Algorithms}
\label{Algo1}
\begin{algorithmic}[1]
\STATE \textbf{Input:}   \( \text{x} \), Model \targetmodel, Model \referencemodel, threshold \threshold.
\STATE \textbf{Output:} Class of x

\STATE \textbf{Steps:}
\STATE \(x_{masked}\)=\( x \) with one unit of information masked.
\STATE Use Model \targetmodel and Model \referencemodel to predict/generate data from  \(x_{masked}\). \\ $\hat{y}_t$ = \targetmodel (\(x_{masked}\)) and $\hat{y}_r$ = \referencemodel (\(x_{masked}\))
\STATE Calculate the loss \( L_T \) between the predicted/generated data by Model \targetmodel  and the original data.  \( L_T(x) \)=${Loss}_{\text{DTW}}(\hat{y}_t, x) $ 
\STATE Calculate the loss \( L_R(x) \) between the predicted/generated data by Model \referencemodel   and the original data. \( L_R (x)\)=${Loss}_{\text{DTW}}(\hat{y}_r, x) $
\STATE Compute the ratio \( R(x) = \frac{L_T(x)}{L_R(x)} \).
\STATE Return 1 if  \( R \leq \theta \) else return 0.
\end{algorithmic}
\end{algorithm}

\subsection{Constructing the Reference Model for Benchmarking}
One of the key innovations introduced is the concept of a reference model, named as Model \referencemodel. The primary function of Model \referencemodel is to serve as a benchmark for comparing the outputs of Model \targetmodel with those of an unbiased model. This comparison helps to determine whether Model \targetmodel is making accurate predictions or just memorizing the training data. In particular, if both models have similar outputs, it indicates that the prediction is likely to be genuine. Conversely, a significant discrepancy between the predictions indicate a high probability that the data point $x$ is memorized and comes from the training set of the Model \targetmodel.

To ensure the effectiveness of the attack, it is essential that Model \targetmodel and Model \referencemodel have equivalent performance. One method to validate this equivalence is to evaluate the performance of the models on publicly available datasets.

The construction of Model \referencemodel depends on the specific attack scenario. The closer Model \referencemodel is to Model \targetmodel, the greater the likelihood of a successful attack.

The attacker can use a publicly available time series imputation model for Model \referencemodel. This method uses existing, well-validated models as a reference, ensuring that Model \referencemodel provides an unbiased benchmark for comparison.

Alternatively, if the attacker has some meta-information about the target model, they can attempt to train a similar model themselves using a public dataset. It is crucial that the attacker makes sure that the trained model achieves equivalent performance to Model \targetmodel for the attack to succeed.

\subsection{Evaluating Predictions and Determining Membership}\label{R-score}
After defining Model \referencemodel, we proceed to the prediction and evaluation phase to compute the ratio $R$. This phase involves processing of the predictions from both Model \targetmodel and Model \referencemodel and comparing their outputs to determine the reliability of the predictions.

For  data $x$, the predicted outputs by Model \targetmodel and Model \referencemodel are denoted as $\hat{y}_t$ and $\hat{y}_r$, respectively. These predictions are then used to calculate the loss functions for both models.

 The loss function for the attack is different from the one used to train the models. In our case, the loss function to compute $L_T(x)$ and $L_R(x)$ is the Dynamic Time Warping (DTW) similarity function. The DTW similarity  function measures the similarity between two time series by aligning them in such a way that the distance between the corresponding points is minimized. The  DTW loss formula is given by the following equation:
\[ \text{Loss}_{\text{DTW}}(y_1, y_2) = \min \left( \sum_{i=1}^{N} \sum_{j=1}^{M} d(y_1(i), y_2(j)) \right) \]
where $d(y_1(i), y_2(j))$ is the distance between the $i$-th point of $y_1$ and the $j$-th point of $y_2$, and the minimization is taken over all possible alignments of the two time series.\\ 

The ratio $R$ of the losses is computed as follows:
\[ R (x)= \frac{L_T(x)}{L_R(x)} \]

A threshold value $\theta$ is defined to classify the data:
\[ \text{Classify } x \text{ as a member of the training data if } R(x) \leq \theta \]

\subsection{Determining the Threshold Value \threshold }
This threshold $\theta$ is a critical parameter that helps distinguish between genuine predictions and memorized data. If the ratio $R(x)$ exceeds the threshold, it indicates that the data point $x$ is likely to be part of the training set, suggesting that Model \targetmodel has memorized this data point rather than making a genuine prediction.

There are several ways to determine the threshold (\threshold). One approach is to take a set of data that the malicious user knows is not used by the target model (\targetmodel) and compute the mean of the score $R$ of the test set. Then, define \threshold, as the mean of the test data plus n times the standard deviation:
\[\theta = \text{mean}(R(x_{test})) + n \times \text{std}(R(x_{test})) \]

Another approach is to take the top $n\%$ of the \( R(x) \) score. This method can be nuanced by the fact that the attacker must have an approximate knowledge of the number of data points belonging to the dataset.

\section{Performance Evaluation} %Experimentation}
To evaluate the effectiveness of the LBRM approach, we focused on black-box API attacks. We conducted experiments on two different types of architectures: a predictive model, SAITS \cite{du_saits_2023}, and a generative model, AE \cite{fu_filling_2024}. We tested the LBRM approach in two scenarios: one without fine-tuning and one with fine-tuning. By evaluating these scenarios, we aim to demonstrate the versatility and robustness of the LBRM approach in different contexts.

\subsection{Experimental Methodology}
To evaluate our attack algorithm, we use two time series imputation architectures: SAITS \cite{du_saits_2023}, which is predictive, and an autoencoder \cite{fu_filling_2024}, which is generative. We chose these architectures to take advantage of their different approaches to handling missing values.\\

\subsubsection{Metrics.} We use the same metrics as in \cite{carlini_mia_2022,liu_membership_2022} to summarize the risk and probability of success of the attack. The evaluation metrics are:

\begin{itemize}
\item \textbf{TPR at Low FPR.} Specifies the true positive rate when the false positive rate is fixed at 0.1, highlighting the precision of the attack under critical conditions.
 \item \textbf{ROC Curve.} Compares the ratio of true positives to false positives at different thresholds, providing a visual representation of the model's performance.
\item \textbf{AUROC Score.} The AUROC is the area under the ROC curves. Measures the ability of the model to discriminate between members and non-members of the training set across all thresholds. 

\end{itemize}

\subsubsection{Datasets.} We use two datasets of electric consumption patterns: the London SmartMeter Energy Consumption Data \cite{noauthor_smartmeter_nodate} to evaluate SAITS \cite{du_saits_2023} and the ASHRAE dataset \cite{noauthor_ashrae_nodate} to evaluate the autoencoder \cite{fu_filling_2024}. 

\begin{itemize}
    \item London smart SmartMeter Energy Consumption Data (LSMEC) \cite{noauthor_smartmeter_nodate} is a dataset that provides time series of energy consumption. Energy consumption readings were collected from 5,567 London households as part of the Low Carbon London project (Nov 2011 - Feb 2014). Readings were taken every half hour. The dataset includes energy consumption (kWh per half hour), household ID, date, and time, for a total of approximately 167 million rows.
    \item   ASHRAE dataset \cite{noauthor_ashrae_nodate} This dataset contains one year of hourly which is about 8784 point meter readings from 1479 buildings in various locations around the world. It includes energy consumption readings for four energy types: electricity, chilled water, steam, and hot water. The dataset also includes building metadata and weather data to support the development of counterfactual models for assessing energy efficiency improvements.
\end{itemize}

In our evaluation, we compare the results obtained using our algorithm (LBRM) with those obtained using a naive approach based on loss alone. The main difference between our approach (LBRM) and the naive loss approach lies in the way the data is separated. As explained in Section \ref{R-score}, our main contribution is to use the $R(x)$ score to process membership, whereas in the naive loss approach, we  only use $L_T(x)$ to discriminate between training and test data.

%IVAN: Mieux vendre l'intérêt de la comparaison: il s'agit de montrer que c'est bien le modèle témoin qui permet la réussite de l'attaque ok.
%Rajouter le descriptif de l'attaque: on prend Lp(x) au lieu de R(x).
\subsubsection{Naive loss attack.} For benchmarking purposes, we conducted what we refer to as a Naive Loss Attack. This attack aims to separate training data from test data based solely on the loss, similar to the work presented in \cite{liu_membership_2022}. The underlying principle of this attack is that models tend to have lower loss on training data compared to test data, because they have been optimized specifically on the training set. By exploiting this difference in loss, we can infer whether a particular data point was part of the training set or not.

%IVAN Je ne suis pas sûr que threat model soit le plus adapté, on pourrait juste dire training scenario ? ok
\subsubsection{Experiment Scenarios.} In our threat model, we consider two scenarios: one without fine-tuning and one with fine-tuning. Since our approach requires a reference model (\referencemodel) as input, we consider two realistic scenarios. In the first scenario, the target model is not fine-tuned and is trained exclusively on private data. In the second scenario, the target model is first trained on public data and then fine-tuned on private data. These two scenarios capture the different contexts in which the target model might operate.\\

We consider a target model \targetmodel and two datasets: a public dataset \publicdataset  and a private dataset \privatedataset. We analyze two scenarios:

\paragraph{Scenario 1 (unfine-tuned):} 
The target model \targetmodel is trained only on the private dataset (\privatedataset). We make sure that \privatedataset and \publicdataset are from different distributions. This distinction is critical to demonstrating that the attack does not require access to data similar to that of the model \targetmodel .

\paragraph{Scenario 2 (fine-tuned):} 
The target model \targetmodel is first trained on the public dataset (\publicdataset) and then fine-tuned on the private dataset (\privatedataset). 
\\

In both scenarios, the model is accessible via an API for imputing missing values, and a malicious user wants to extract private data using our LBRM algorithm, having access to the public dataset and some knowledge of the model's architecture.

\subsubsection{Training models \targetmodel and \referencemodel.}
For experimental purposes, we assume that the attacker has access to the meta-parameters and trains their model using the same methodologies. However, it is important to note that the attack can still be effective even if the models are based on different architectures. The only requirement is that the reference model and the target model have comparable performance levels. Both the target model (\targetmodel) and the reference model (\referencemodel) are trained using the SAITS and Autoencoder architectures. The architectural details for both models are outlined below.

\paragraph{SAITS Architecture:}
The SAITS architecture is configured with a single feature (electricity consumption). It comprises two layers, each with a model dimension of 512. The feed-forward layers have a dimension of 128, and the model employs four attention heads, each with key and value dimensions of 128. No dropout is applied. The model is trained for 100 epochs on a GPU, with a batch size of 64.

\paragraph{Autoencoder Architecture:}
The Autoencoder architecture is designed to handle temporal data. The model comprises an encoder with two layers and a decoder with two layers, both connected through a fully-connected layer at the bottleneck. Similar to the SAITS architecture, it is trained for 100 epochs on a GPU without dropout.\\

\paragraph{Data preparation:} each dataset is divided randomly into public, private, and test sets. The models \targetmodel are trained on the public dataset \publicdataset combined with the private dataset \privatedataset. The models \referencemodel are trained solely on the public dataset \publicdataset. In Table \ref{tab:data_preparation}, we summarize the division of each dataset. The division ratios are as follows:
\begin{itemize}
\item \textit{Scenario 1}: In this scenario, we aim to train the \referencemodel on a different data distribution than the \targetmodel. For this purpose, we randomly divide the data into 2/5 for the \publicdataset and 2/5 for the \privatedataset, and reserve 1/5 for testing. The division for each dataset is as follows: 
\begin{itemize} 
    \item \textbf{LSMEC}: Both the \publicdataset and \privatedataset contain 2226 time series each, while the test dataset has 1113 time series. 
    \item \textbf{ASHRAE}: Both the \publicdataset and \privatedataset contain 590 time series each, while the test dataset has 295 time series. 
\end{itemize}

\item \textit{Scenario 2}: In this scenario, we consider the case of fine-tuning. For this purpose, we divide the data such that 3/5 is used for the \publicdataset, 1/5 for the \privatedataset, and 1/5 for the test dataset. The division for each dataset is as follows:
\begin{itemize}
    \item \textbf{LSMEC}: The \publicdataset contains 3340 time series, while the \privatedataset and test datasets each contain 1113 time series.
    \item \textbf{ASHRAE}: The \publicdataset contains 887 time series, while the \privatedataset and test datasets each contain 295 time series.
\end{itemize}
\end{itemize}

\begin{table}[H]
\centering
\resizebox{1\columnwidth}{!}{%
\begin{tabular}{@{}ccccc@{}}
\toprule
 & \multicolumn{2}{c}{Scenario 1} & \multicolumn{2}{c}{Scenario 2 (fine tuning)} \\ \midrule
\multicolumn{1}{c|}{} & \multicolumn{1}{c|}{LSMEC (SAITS)} & \multicolumn{1}{c|}{ASHRAE (AE)} & \multicolumn{1}{c|}{LSMEC (SAITS)} & ASHRAE (AE) \\ \midrule
\multicolumn{1}{c|}{\textit{\textbf{Public} \publicdataset}} & 2226 & \multicolumn{1}{c|}{590} & 3340 & 887 \\ \midrule
\multicolumn{1}{c|}{\textit{\textbf{Private} \privatedataset}} & 2226 & \multicolumn{1}{c|}{590} & 1113 & 295 \\ \midrule
\multicolumn{1}{c|}{\textit{\textbf{Test}}} & 1113 & \multicolumn{1}{c|}{295} & 1113 & 295 \\ \bottomrule
\end{tabular}%
}
\caption{Division of datasets in different scenarios}
\label{tab:data_preparation}
\end{table}

\subsection{Verifying the Validity of the Models}

%IVAN trouver un autre mot que result: properties, performance ? 
%\subsubsection{Training models performance.}
One input to the LBRM algorithm is a reference model, model \referencemodel. To maximize the success of the attack, we must ensure that the model \referencemodel has similar performance to the model \targetmodel. This similarity in performance is crucial because the effectiveness of the attack is highly sensitive to the performance parity between the reference and target models. If there is a significant discrepancy in their performance, the success rate of the attack may be adversely affected.

In addition, to minimize bias in our experiments, we must ensure that model \targetmodel does not have significant issues with overfitting. This precaution helps maintain the integrity of our experimental results and ensures that the observed outcomes are due to the attack methodology rather than artifacts of model training.

To evaluate the performance of each model, we randomly remove 20\% of the data and test the resulting model on that data by calculating the Mean Absolute Error (MAE). The results are shown in Table \ref{tab:trainin_models}.

\begin{table}[H]
\centering
\resizebox{\columnwidth}{!}{%
\begin{tabular}{@{}ccccccc@{}}
\toprule
 & \multicolumn{2}{c}{Scenario 1} & \multicolumn{4}{c}{Scenario 2 (fine tuning)} \\ \midrule
\multicolumn{1}{c|}{} & \multicolumn{1}{c|}{SAITS} & \multicolumn{1}{c|}{AE} & \multicolumn{2}{c|}{SAITS} & \multicolumn{2}{c}{AE} \\ \midrule
\multicolumn{1}{c|}{\textit{\textbf{MAE}}} & \multicolumn{1}{c|}{Model \targetmodel} & \multicolumn{1}{c|}{Model \targetmodel} & Model \targetmodel & \multicolumn{1}{c|}{Model \referencemodel} & Model \targetmodel & Model \referencemodel \\ \midrule
\multicolumn{1}{c|}{Data Train} & \multicolumn{1}{c|}{0.175} & \multicolumn{1}{c|}{0.15} & 0.19 & \multicolumn{1}{c|}{0.185} & 0.019 & 0.018 \\
\multicolumn{1}{c|}{Data Test} & \multicolumn{1}{c|}{0.20} & \multicolumn{1}{c|}{0.2} & 0.24 & \multicolumn{1}{c|}{0.21} & 0.024 & 0.021 \\ \bottomrule
\end{tabular}%
}
\caption{Performance Evaluation of SAITS and AE Models Based on Mean Absolute Error (MAE)}
\label{tab:trainin_models}
\end{table}

Table \ref{tab:trainin_models} shows the performance evaluation of the SAITS and AE models based on Mean Absolute Error (MAE) on the two scenarios. The evaluation involves randomly removing 20\% of the data and testing the resulting models on that data.

The SAITS model and the AE model both demonstrated training and test MAE values that were very close to each other. For instance, the difference between their training and test MAE values was only 0.025 in \textit{Scenario 1}. These small gaps indicate minimal overfitting which apply low risk vulnerability to naive loss and confirm that the performance of the reference model Model \referencemodel and the target model \targetmodel are quite similar.

\subsection{Experiment Results of the LBRM Approcah to detect MIA  Attacks }

%\subsubsection{Attack result.}

Table \ref{tab:LBRM_result} presents a comparative analysis of attack success rates using our algorithm (LBRM) and Naive Loss for two models, SAITS and AE, over the two scenarios. 

\begin{table}[H]
\centering
\resizebox{\columnwidth}{!}{%
\begin{tabular}{@{}ccccc|cccc@{}}
\toprule
 & \multicolumn{4}{c|}{\textit{Scenario 1}} & \multicolumn{4}{c}{\textit{Scenario 2 (fine tuning)}} \\ \midrule
\multicolumn{1}{c|}{} & \multicolumn{2}{c|}{SAITS} & \multicolumn{2}{c|}{AE} & \multicolumn{2}{c|}{SAITS} & \multicolumn{2}{c}{AE} \\ \midrule
\multicolumn{1}{c|}{\textit{\textbf{}}} & \textbf{LBRM} & \multicolumn{1}{c|}{Naive Loss} & \textbf{LBRM} & Naive Loss & \textbf{LBRM} & \multicolumn{1}{c|}{Naive Loss} & \textbf{LBRM} & Naive Loss \\ \midrule
\multicolumn{1}{c|}{AUROC} & \textbf{0.71} & \multicolumn{1}{c|}{0.52} & \textbf{0.77} & 0.42 & \textbf{0.90} & \multicolumn{1}{c|}{0.55} & \textbf{0.85} & 0.52 \\
\multicolumn{1}{c|}{TPR@0.1} & \textbf{0.26} & \multicolumn{1}{c|}{0.11} & \textbf{0.45} & 0.08 & \textbf{0.75} & \multicolumn{1}{c|}{0.12} & \textbf{0.56} & 0.12 \\
\multicolumn{1}{l|}{TPR@top25\%} & \textbf{60\%} & \multicolumn{1}{c|}{52\%} & \textbf{72\%} & 40\% & \textbf{92\%} & \multicolumn{1}{c|}{48\%} & \textbf{80\%} & 51\% \\ \bottomrule
\end{tabular}%
}

\caption{Comparative Analysis of Attack Success Based on AUROC, TPR@0.1  and TPR@top25\% for SAITS and AE Models in Two Scenarios}
\label{tab:LBRM_result}
\end{table}

The LBRM approach consistently outperformed the Naive Loss approach across all scenarios. On average, the LBRM approach significantly improved key performance measures. For example, in Scenario 1, the SAITS model's AUROC increased from 0.52 (close to random performance) to 0.71, while the AE model's AUROC improved from 0.55 (also close to random) to 0.77. In Scenario 2, the AUROC of the SAITS model increases from 0.55 to 0.90, and the AUROC of the AE model increases from 0.52 to 0.85. Additionally, the TPR@0.1 values showed significant improvements, indicating a higher rate of true positive detection.

\subsubsection{Impact of the Threshold Value \threshold.}
A key factor in the effectiveness of the attack is how the threshold (\threshold) is defined, which is an input to the LBRM algorithm. As mentioned in Section \ref{R-score}, there are several ways to define \threshold. Through experimentation, we have found that the risk of the attack is independent of the threshold (\threshold) and that the attack also works with the top n\%  of the \( R(x) \) score scenario.

\begin{figure}[H]
    \centering
    \begin{subfigure}[b]{0.4\textwidth}
        \centering
        \includegraphics[width=\textwidth]{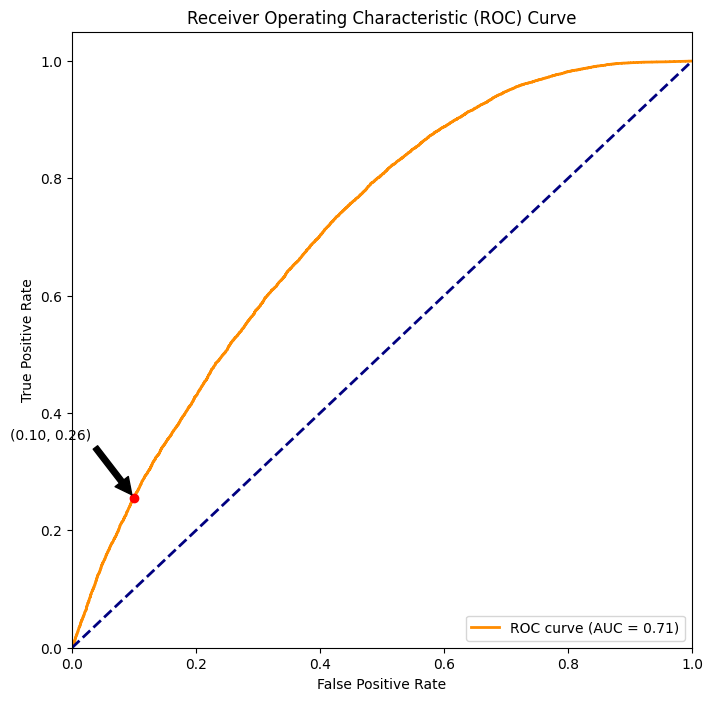}
        \caption{SAITS model}
        \label{fig:figure1}
    \end{subfigure}
    \hfill
    \begin{subfigure}[b]{0.4\textwidth}
        \centering
        \includegraphics[width=\textwidth]{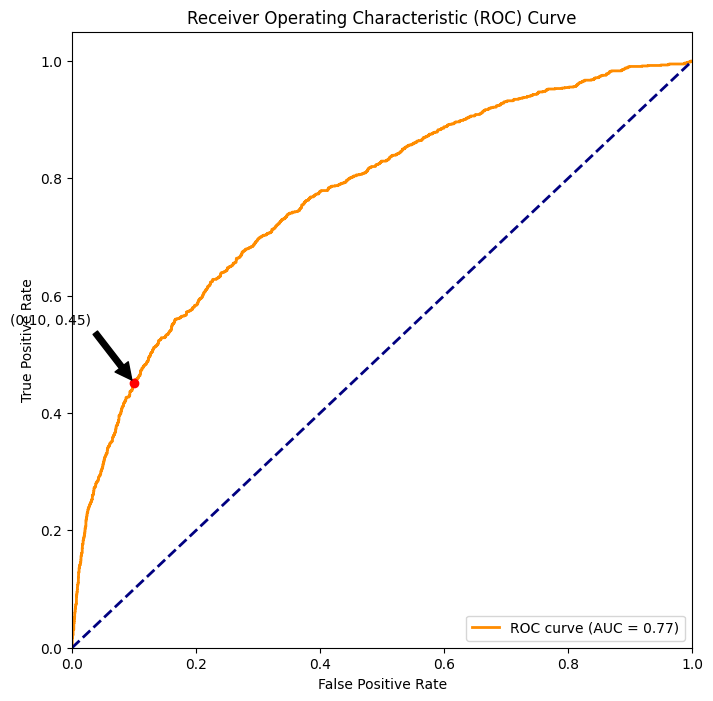}
        \caption{ Autoencoder model}
        \label{fig:figure2}
    \end{subfigure}
    \caption{ Result of the membership attack for scenario 1 }
    \label{fig:roc1}
\end{figure}

\begin{figure}[H]
    \centering
    \begin{subfigure}[b]{0.4\textwidth}
        \centering
        \includegraphics[width=\textwidth]{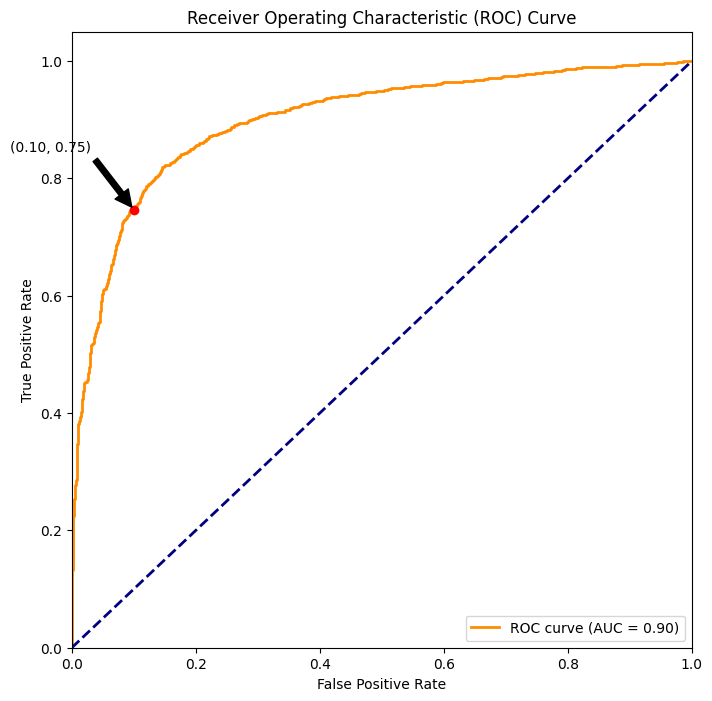}
        \caption{SAITS model}
        \label{fig:figure1}
    \end{subfigure}
    \hfill
    \begin{subfigure}[b]{0.4\textwidth}
        \centering
        \includegraphics[width=\textwidth]{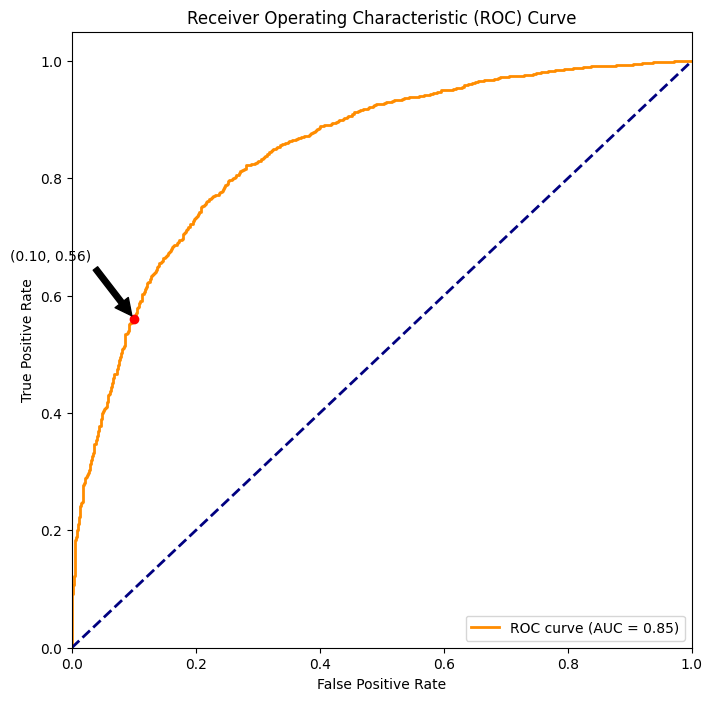}
        \caption{ Autoencoder model}
        \label{fig:figure2}
    \end{subfigure}
    \caption{ Result of the membership attack for scenario 2 (fine tuning)}
    \label{fig:roc2}
\end{figure}

In Figure \ref{fig:roc1} and \ref{fig:roc2}, we observe the results of our algorithms (LBRM) applied to the autoencoder architecture and SAITS in the context of scenario 1 and 2. The ROC curve shows that, regardless of how we choose the threshold \threshold, the risk remains high due to the shape of the curve, which is wider than 0.5. For example, the optimal point for the autoencoder architecture has a true positive rate (TPR) of 0.56 and a false positive rate of 0.1. This represents a favorable trade-off for the attacker, making the model vulnerable to membership inference attacks and potentially other types of attacks. And for SAITS, the optimal point achieves a TPR of 0.75 with a false positive rate of 0.1.

Another approach is to take the top $n\%$ of the \( R(x) \) score, as shown in Table \ref{tab:LBRM_result}. For our experiments, we set the $n=25\%$ because in our experiments we know we need to evaluate 50\% as private data and 50\% as test, Table \ref{tab:LBRM_result} shows the true positive rate (TPR) of the top 25\% \( R(x) \) score for each scenario, highlighting the effectiveness of the attack across different architectures and scenarios. The high TPR values indicate a significant risk, as the attack remains effective even when targeting the top 25\% of \( R(x) \) scores.

\section{Conclusion}
The LBRM (Loss-Based with Reference Model) approach significantly enhances the accuracy of membership inference attacks by utilizing a reference model to distinguish between training and test data. This method is particularly effective in scenarios where the target model is either trained exclusively on private data or fine-tuned on private data after initial training on public data.

Our experimental results demonstrate that the LBRM approach consistently outperforms the naive loss approach, providing substantial improvements in key performance metrics such as AUROC and TPR. Specifically, in Scenario 1 (without fine tuning), the SAITS model AUROC increased by 36.5\%, and the AE model  AUROC improved by 40\% . In Scenario 2 (with fine tuning), the SAITS model AUROC increased by 63.6\%, and the AE model  AUROC improved by 63.5\% .

The versatility of the LBRM method across different scenarios and architectures, including transformers and autoencoders, underscores its robustness and reliability. Despite its complexity and dependency on the reference model, the LBRM approach offers a powerful tool for detecting subtle memorization patterns, thereby addressing privacy risks in time series imputation models. Future work may explore further optimization of the reference model and the application of LBRM to other types of generative models.

\bibliographystyle{unsrtnat}
\bibliography{bib}

\begin{thebibliography}{23}
\providecommand{\natexlab}[1]{#1}
\providecommand{\url}[1]{\texttt{#1}}
\expandafter\ifx\csname urlstyle\endcsname\relax
  \providecommand{\doi}[1]{doi: #1}\else
  \providecommand{\doi}{doi: \begingroup \urlstyle{rm}\Url}\fi

\bibitem[Carlini et~al.()Carlini, Tramèr, Wallace, Jagielski, Herbert-Voss, Lee, Roberts, Brown, Song, Erlingsson, Oprea, and Raffel]{carlini_extracting_nodate}
Nicholas Carlini, Florian Tramèr, Eric Wallace, Matthew Jagielski, Ariel Herbert-Voss, Katherine Lee, Adam Roberts, Tom Brown, Dawn Song, Úlfar Erlingsson, Alina Oprea, and Colin Raffel.
\newblock Extracting {Training} {Data} from {Large} {Language} {Models}.

\bibitem[Carlini et~al.(2019)Carlini, Liu, Erlingsson, Kos, and Song]{carlini_secret_2019}
Nicholas Carlini, Chang Liu, Úlfar Erlingsson, Jernej Kos, and Dawn Song.
\newblock The {Secret} {Sharer}: {Evaluating} and {Testing} {Unintended} {Memorization} in {Neural} {Networks}, July 2019.
\newblock URL \url{http://arxiv.org/abs/1802.08232}.
\newblock arXiv:1802.08232.

\bibitem[Liu et~al.(2021)Liu, Jia, Qu, and Gong]{liu_encodermi_2021}
Hongbin Liu, Jinyuan Jia, Wenjie Qu, and Neil~Zhenqiang Gong.
\newblock {EncoderMI}: {Membership} {Inference} against {Pre}-trained {Encoders} in {Contrastive} {Learning}.
\newblock In \emph{Proceedings of the 2021 {ACM} {SIGSAC} {Conference} on {Computer} and {Communications} {Security}}, {CCS} '21, pages 2081--2095, New York, NY, USA, November 2021. Association for Computing Machinery.
\newblock ISBN 978-1-4503-8454-4.
\newblock \doi{10.1145/3460120.3484749}.
\newblock URL \url{https://dl.acm.org/doi/10.1145/3460120.3484749}.

\bibitem[Fu et~al.(2023)Fu, Wang, Gao, Liu, Li, and Jiang]{fu_practical_2023}
Wenjie Fu, Huandong Wang, Chen Gao, Guanghua Liu, Yong Li, and Tao Jiang.
\newblock Practical {Membership} {Inference} {Attacks} against {Fine}-tuned {Large} {Language} {Models} via {Self}-prompt {Calibration}, December 2023.
\newblock URL \url{http://arxiv.org/abs/2311.06062}.
\newblock arXiv:2311.06062 [cs].

\bibitem[Wu et~al.(2021)Wu, Yang, Pan, and Yuan]{wu_adapting_2021}
Bang Wu, Xiangwen Yang, Shirui Pan, and Xingliang Yuan.
\newblock Adapting {Membership} {Inference} {Attacks} to {GNN} for {Graph} {Classification}: {Approaches} and {Implications}, October 2021.
\newblock URL \url{http://arxiv.org/abs/2110.08760}.
\newblock arXiv:2110.08760 [cs].

\bibitem[He et~al.(2021)He, Wen, Wu, Backes, Shen, and Zhang]{he_node-level_2021}
Xinlei He, Rui Wen, Yixin Wu, Michael Backes, Yun Shen, and Yang Zhang.
\newblock Node-{Level} {Membership} {Inference} {Attacks} {Against} {Graph} {Neural} {Networks}, February 2021.
\newblock URL \url{http://arxiv.org/abs/2102.05429}.
\newblock arXiv:2102.05429 [cs].

\bibitem[Carlini et~al.(2022{\natexlab{a}})Carlini, Chien, Nasr, Song, Terzis, and Tramer]{carlini_membership_2022}
Nicholas Carlini, Steve Chien, Milad Nasr, Shuang Song, Andreas Terzis, and Florian Tramer.
\newblock Membership {Inference} {Attacks} {From} {First} {Principles}, April 2022{\natexlab{a}}.
\newblock URL \url{http://arxiv.org/abs/2112.03570}.
\newblock arXiv:2112.03570.

\bibitem[Webster et~al.(2021)Webster, Rabin, Simon, and Jurie]{webster_this_2021}
Ryan Webster, Julien Rabin, Loic Simon, and Frederic Jurie.
\newblock This {Person} ({Probably}) {Exists}. {Identity} {Membership} {Attacks} {Against} {GAN} {Generated} {Faces}, July 2021.
\newblock URL \url{http://arxiv.org/abs/2107.06018}.
\newblock arXiv:2107.06018.

\bibitem[Olatunji et~al.(2021)Olatunji, Nejdl, and Khosla]{olatunji_membership_2021}
Iyiola~E. Olatunji, Wolfgang Nejdl, and Megha Khosla.
\newblock Membership {Inference} {Attack} on {Graph} {Neural} {Networks}, December 2021.
\newblock URL \url{http://arxiv.org/abs/2101.06570}.
\newblock arXiv:2101.06570.

\bibitem[Zhang et~al.(2024)Zhang, Yu, Wen, Backes, and Zhang]{zhang_generated_2024}
Minxing Zhang, Ning Yu, Rui Wen, Michael Backes, and Yang Zhang.
\newblock Generated {Distributions} {Are} {All} {You} {Need} for {Membership} {Inference} {Attacks} {Against} {Generative} {Models}.
\newblock pages 4839--4849, 2024.

\bibitem[Wang et~al.(2025)Wang, Du, Yang, Qian, Cao, Zhang, Wang, Liang, and Wen]{wang_deep_2025}
Jun Wang, Wenjie Du, Yiyuan Yang, Linglong Qian, Wei Cao, Keli Zhang, Wenjia Wang, Yuxuan Liang, and Qingsong Wen.
\newblock Deep {Learning} for {Multivariate} {Time} {Series} {Imputation}: {A} {Survey}, February 2025.
\newblock URL \url{http://arxiv.org/abs/2402.04059}.
\newblock arXiv:2402.04059 [cs].

\bibitem[et~al(2024)]{openai_gpt-4_2024}
OpenAI et~al.
\newblock {GPT}-4 {Technical} {Report}, March 2024.
\newblock URL \url{http://arxiv.org/abs/2303.08774}.
\newblock arXiv:2303.08774.

\bibitem[Touvron et~al.(2023)Touvron, Lavril, Izacard, Martinet, Lachaux, Lacroix, Rozière, Goyal, Hambro, Azhar, Rodriguez, Joulin, Grave, and Lample]{touvron_llama_2023}
Hugo Touvron, Thibaut Lavril, Gautier Izacard, Xavier Martinet, Marie-Anne Lachaux, Timothée Lacroix, Baptiste Rozière, Naman Goyal, Eric Hambro, Faisal Azhar, Aurelien Rodriguez, Armand Joulin, Edouard Grave, and Guillaume Lample.
\newblock {LLaMA}: {Open} and {Efficient} {Foundation} {Language} {Models}, February 2023.
\newblock URL \url{http://arxiv.org/abs/2302.13971}.
\newblock arXiv:2302.13971.

\bibitem[Jiang et~al.(2023)Jiang, Sablayrolles, Mensch, Bamford, Chaplot, Casas, Bressand, Lengyel, Lample, Saulnier, Lavaud, Lachaux, Stock, Scao, Lavril, Wang, Lacroix, and Sayed]{jiang_mistral_2023}
Albert~Q. Jiang, Alexandre Sablayrolles, Arthur Mensch, Chris Bamford, Devendra~Singh Chaplot, Diego de~las Casas, Florian Bressand, Gianna Lengyel, Guillaume Lample, Lucile Saulnier, Lélio~Renard Lavaud, Marie-Anne Lachaux, Pierre Stock, Teven~Le Scao, Thibaut Lavril, Thomas Wang, Timothée Lacroix, and William~El Sayed.
\newblock Mistral {7B}, October 2023.
\newblock URL \url{http://arxiv.org/abs/2310.06825}.
\newblock arXiv:2310.06825.

\bibitem[Liu et~al.(2022)Liu, Zhao, Backes, and Zhang]{liu_membership_2022}
Yiyong Liu, Zhengyu Zhao, Michael Backes, and Yang Zhang.
\newblock Membership {Inference} {Attacks} by {Exploiting} {Loss} {Trajectory}, August 2022.
\newblock URL \url{http://arxiv.org/abs/2208.14933}.
\newblock arXiv:2208.14933 [cs].

\bibitem[Hilprecht et~al.(2019{\natexlab{a}})Hilprecht, Härterich, and Bernau]{hilprecht_monte_2019}
Benjamin Hilprecht, Martin Härterich, and Daniel Bernau.
\newblock Monte {Carlo} and {Reconstruction} {Membership} {Inference} {Attacks} against {Generative} {Models}.
\newblock volume 2019, July 2019{\natexlab{a}}.
\newblock \doi{10.2478/popets-2019-0067}.

\bibitem[Hilprecht et~al.(2019{\natexlab{b}})Hilprecht, Härterich, and Bernau]{hilprecht_reconstruction_2019}
Benjamin Hilprecht, Martin Härterich, and Daniel Bernau.
\newblock Reconstruction and {Membership} {Inference} {Attacks} against {Generative} {Models}, June 2019{\natexlab{b}}.
\newblock URL \url{http://arxiv.org/abs/1906.03006}.
\newblock arXiv:1906.03006 [cs].

\bibitem[Hayes et~al.(2018)Hayes, Melis, Danezis, and Cristofaro]{hayes_logan_2018}
Jamie Hayes, Luca Melis, George Danezis, and Emiliano~De Cristofaro.
\newblock {LOGAN}: {Membership} {Inference} {Attacks} {Against} {Generative} {Models}, August 2018.
\newblock URL \url{http://arxiv.org/abs/1705.07663}.
\newblock arXiv:1705.07663.

\bibitem[Du et~al.(2023)Du, Cote, and Liu]{du_saits_2023}
Wenjie Du, David Cote, and Yan Liu.
\newblock {SAITS}: {Self}-{Attention}-based {Imputation} for {Time} {Series}.
\newblock \emph{Expert Systems with Applications}, 219:\penalty0 119619, June 2023.
\newblock ISSN 09574174.
\newblock \doi{10.1016/j.eswa.2023.119619}.
\newblock URL \url{http://arxiv.org/abs/2202.08516}.
\newblock arXiv:2202.08516 [cs].

\bibitem[Fu et~al.(2024)Fu, Quintana, Nagy, and Miller]{fu_filling_2024}
Chun Fu, Matias Quintana, Zoltan Nagy, and Clayton Miller.
\newblock Filling time-series gaps using image techniques: {Multidimensional} context autoencoder approach for building energy data imputation.
\newblock \emph{Applied Thermal Engineering}, 236:\penalty0 121545, January 2024.
\newblock ISSN 1359-4311.
\newblock \doi{10.1016/j.applthermaleng.2023.121545}.
\newblock URL \url{https://www.sciencedirect.com/science/article/pii/S1359431123015740}.

\bibitem[Carlini et~al.(2022{\natexlab{b}})Carlini, Chien, Nasr, Song, Terzis, and Tramèr]{carlini_mia_2022}
Nicholas Carlini, Steve Chien, Milad Nasr, Shuang Song, Andreas Terzis, and Florian Tramèr.
\newblock Membership inference attacks from first principles.
\newblock In \emph{2022 IEEE Symposium on Security and Privacy (SP)}, pages 1897--1914, 2022{\natexlab{b}}.
\newblock \doi{10.1109/SP46214.2022.9833649}.

\bibitem[noa({\natexlab{a}})]{noauthor_smartmeter_nodate}
{SmartMeter} {Energy} {Consumption} {Data} in {London} {Households} - {London} {Datastore}, {\natexlab{a}}.
\newblock URL \url{https://data.london.gov.uk/dataset/smartmeter-energy-use-data-in-london-households}.

\bibitem[noa({\natexlab{b}})]{noauthor_ashrae_nodate}
{ASHRAE} - {Great} {Energy} {Predictor} {III}, {\natexlab{b}}.
\newblock URL \url{https://kaggle.com/competitions/ashrae-energy-prediction}.

\end{thebibliography}

\end{document}